\pdfoutput=1

\documentclass[11pt]{article}

\usepackage[preprint]{acl}

\usepackage{times}
\usepackage{latexsym}

\usepackage[T1]{fontenc}

\usepackage[utf8]{inputenc}

\usepackage{microtype}

\usepackage{inconsolata}

\usepackage{graphicx}
\usepackage{booktabs} 
\usepackage{array}
\usepackage{multirow}
\usepackage{tabularx} 
\usepackage{siunitx}
\usepackage{subcaption}

\usepackage{amsmath}
\usepackage{amssymb}
\usepackage{mathtools}
\usepackage{amsthm}

\usepackage{algorithm}
\usepackage{algorithmic}

\usepackage{enumitem}

\title{LCD: Advancing Extreme Low-Bit Clustering\\ for Large Language Models via Knowledge Distillation}

\author{Fangxin Liu$^{1,2}$, Ning Yang$^{1,2}$, Junping Zhao$^{3}$, Tao Yang$^{4}$, Haibing Guan$^{1}$ and Li Jiang*$^{1,2}$\\
1.Shanghai Jiao Tong University \quad  2. Shanghai Qi Zhi Institute\\
3.Alibaba Group \quad 4.Huawei Technologies Ltd.\\
\texttt{\{liufangxin, yn937391832, ljiang\_cs\}@sjtu.edu.cn}
}

\begin{document}
\maketitle
\begin{abstract}
Large language models (LLMs) have achieved significant progress in natural language processing but face challenges in deployment due to high memory and computational requirements. Weight quantization is a common approach to address these issues, yet achieving effective low-bit compression remains challenging.
This paper presents LCD, which unifies the learning
of clustering-based quantization within a knowledge distillation framework. Using carefully designed optimization techniques, LCD preserves LLM performance even at ultra-low bit widths of 2–3 bits. Additionally, LCD compresses activations through smoothing and accelerates inference with a LUT-based design. Experimental results show that LCD outperforms existing methods and delivers up to a $6.2\times$ speedup in inference. Notably, LCD is shown to be more cost-effective, making it a practical solution for real-world applications.
\end{abstract}

\section{Introduction}

Large Language Models (LLMs) have demonstrated remarkable capabilities across diverse tasks. However, such escalating model size poses significant challenges in deployment, particularly on resource-constrained devices, due to the substantial memory footprint and computational requirements.

\begin{figure}[t!] 
\vspace{-10pt}
\setlength{\abovecaptionskip}{3pt}
\setlength{\belowcaptionskip}{3pt}
\centering
\includegraphics[width=0.9 \linewidth]{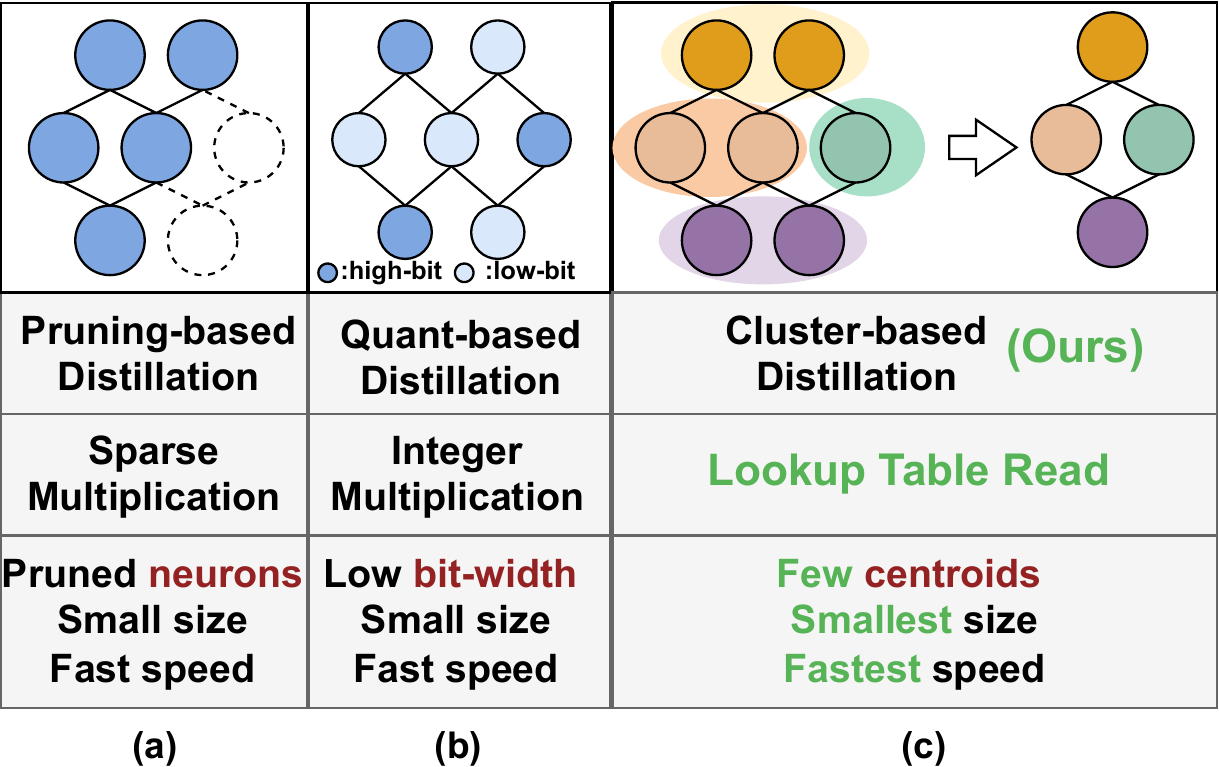}
\caption{Comparison of different distillation methods: (a) Pruning-based: identify and prune unimportant weights; (b) Quantization-based: update weights and reduce bit width; (c) Cluster-based (ours): fine-tune centroid values and counts, enabling LUT instead of multiplication.
}
\label{fig:method-compare}
\vspace{-10pt}
\end{figure}

Weight quantization has emerged as a popular strategy that effectively reduces the model size by representing floating-point numbers with fewer bits~\cite{gholami2022survey}. Recently, 8-bit and even 4-bit quantization techniques have been successfully applied to reduce memory usage and computational overhead while maintaining model performance~\cite{dettmers2023case, frantar2022gptq}. However, quantization lower than 4 bits significantly degrades the fidelity of model weights, leading to deteriorated model performance. To mitigate this decline, huge efforts have been taken for extreme compression of LLM weights. However, ensuring accuracy with extremely low-bit quantization remains a problem that has yet to be solved.

Clustering~\cite{ahmed2020k} is a promising method for low-bit weight compression due to its ability to flexibly adjust the number of centroids, enabling better adaptation to the weight distribution compared to fixed bit-width quantization~\cite{lin2024awq}. While clustering can preserve model accuracy even at lower precision, it faces two challenges:
\begin{itemize}
\vspace{-5pt}
    \item \textbf{Accuracy under extreme low-bit clustering:}
Reducing the number of centroids introduces compression errors. Conventional clustering methods, such as k-means, lack effective techniques for optimizing centroid placement, which exacerbates errors as model sizes grow. This limitation constrains the achievable compression rate and impacts accuracy.
    \item \textbf{Efficient inference with clustered weights:}
Although clustering reduces weights to centroid values, these centroids are often stored and processed as high-bit floating-point numbers, limiting computational benefits. Using indices to represent centroids offers a potential solution, but the presence of outliers in the activations of LLMs makes compression challenging, making index-based representation less effective.
\end{itemize}

To address these challenges, we propose \textbf{LCD}, an extreme low-bit clustering framework combined with distillation for LLMs, as illustrated in Figure~\ref{fig:method-compare}. LCD employs a simple yet effective self-distillation strategy, where the full-precision model acts as its own teacher to guide the low-bit student model. Unlike conventional knowledge distillation (KD) methods that focus on reducing data bits, connections, or neurons, LCD leverages a clustering-based quantization with a smoothing mechanism to minimize quantization errors in both weights and activations. This enables preservation of full-precision model capabilities even at ultra-low-bit quantization levels.
Our contributions are as follows:

\begin{itemize}[itemsep=0pt, parsep=0pt, topsep=0pt]
    \item We introduce LCD, which combines clustering with knowledge distillation to optimize centroid representation. This approach maximizes efficiency while significantly reducing memory requirements.
    \item We present a distillation framework with a density-based centroid initialization and progressive optimization strategies. These techniques minimize clustering-induced errors, ensuring negligible performance degradation even with extreme centroid reduction.
    \item We design a smooth table lookup inference execution that converts both activations and weights into low-bit integer indices. By using index pairs to construct LUTs, LCD replaces complex multiplications, leading to significant computational speedups.
    \item We implement and evaluate LCD. Results show that LCD achieves SOTA accuracy on extremely low-bit LLMs, and achieves up to 6.2x end-to-end inference speedup.
\end{itemize}

\section{Background and Motivation}

\subsection{Quantization in LLM}
Post-Training Quantization (PTQ) and Quantization-Aware Training (QAT) are two primary approaches for compressing LLMs. PTQ avoids retraining by directly quantizing pre-trained models, using techniques such as error minimization~\cite{frantar2022gptq}, outlier handling~\cite{lin2024awq}, vector quantization~\cite{van2024gptvq, liu2024vptq}, and mixed-precision strategies~\cite{li2023llm}. While efficient and training-free, PTQ often degrades in accuracy under low-bit settings due to its lack of adaptation.
QAT, by incorporating quantization into training, improves robustness to quantization noise~\cite{liu2023llm, shao2023omniquant}, but requires full training data and incurs substantial overhead.

Activation quantization is further challenged by large outliers, which can skew dynamic ranges. Some methods isolate outliers, while others like SmoothQuant~\cite{xiao2023smoothquant} alleviate the issue by scaling activations based on weight statistics.
Despite these efforts, achieving efficient and accurate low-bit quantization for LLMs remains an open challenge.

\subsection{Weight Clustering in LLM}

\begin{figure*}[htbp] 
\vspace{-15pt}
\setlength{\abovecaptionskip}{3pt}
\setlength{\belowcaptionskip}{3pt}
\centering
\includegraphics[width=0.8 \linewidth]{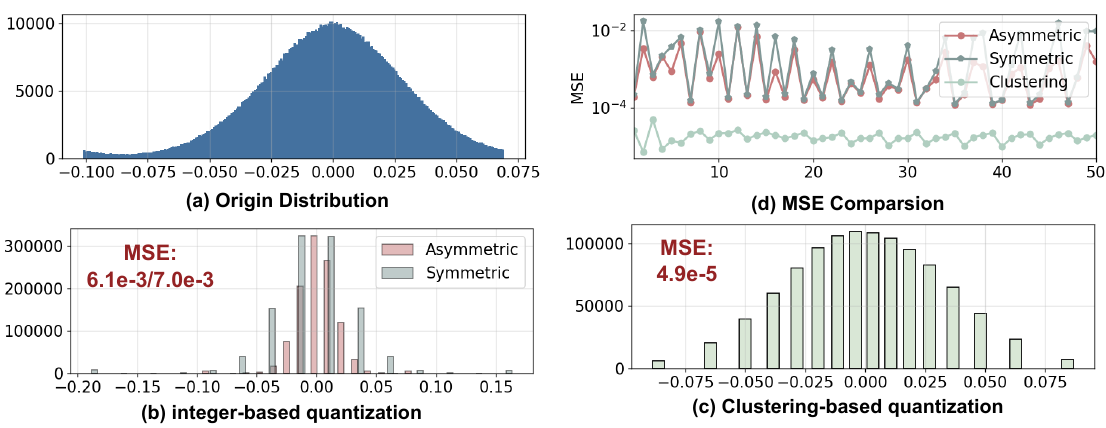}
\vspace{-3pt}
\caption{Comparison of clustering and quantization results and their MSE with same bitwidth (4bits = 16 centroids)}
\label{fig:profiling-clustering}
\vspace{-0.2cm}
\end{figure*}

Clustering~\cite{cho2024edkm} compresses data by grouping similar inputs and representing them with learned centroids. Unlike quantization, which is constrained by fixed bit-widths and predefined codebooks, clustering allows flexible centroid optimization and adaptive control over the number of clusters~\cite{tang2023lut}. This flexibility enables clustering to better approximate the original data distribution, often resulting in lower quantization error. As shown in Figure~\ref{fig:profiling-clustering}, clustering achieves a significantly lower Mean Squared Error (MSE) than conventional quantization. Additionally, clustering-based methods can leverage LUTs for efficient centroid indexing and computation.

Despite these advantages, clustering generally incurs higher computational complexity and storage overhead. Determining the optimal number of centroids is crucial for balancing accuracy and efficiency, yet existing methods often rely on exhaustive search strategies~\cite{liu2024inspire, cho2024edkm}, limiting their scalability in large-scale deployment scenarios.



\subsection{Knowledge Distillation in LLM}

With the increasing size of LLMs, KD has become a key technique for model compression. Initially used for model pruning in DNNs, KD is now commonly combined with quantization to achieve high compression ratios while minimizing accuracy loss.
For example, methods like LLM-QAT~\cite{liu2023llm} and BitDistiller~\cite{du2024bitdistiller} use KD to maintain performance at low bit-widths. Inspired by this, our work integrates KD with clustering to preserve model accuracy while adaptively decreasing centroids.



\subsection{Motivation}
Clustering outperforms quantization in fitting diverse data distributions, offering superior representation capabilities. However, traditional clustering methods like k-means struggle to determine the optimal number of centroids, limiting their ability to achieve extreme compression. Meanwhile, activations are challenging to compress to low bit-widths without accuracy loss, primarily due to outliers.

To overcome these challenges, we propose LCD, a framework that combines clustering with knowledge distillation. LCD achieves precise weight representation with an ultra-low number of centroids and introduces adaptive smoothing to enable low-bit quantization for activations. This dual-side compression significantly enhances acceleration and memory efficiency.


\section{Methodology}

\begin{figure}[t!] 
\setlength{\abovecaptionskip}{3pt}
\setlength{\belowcaptionskip}{3pt}
\centering
\includegraphics[width=1.0 \linewidth]{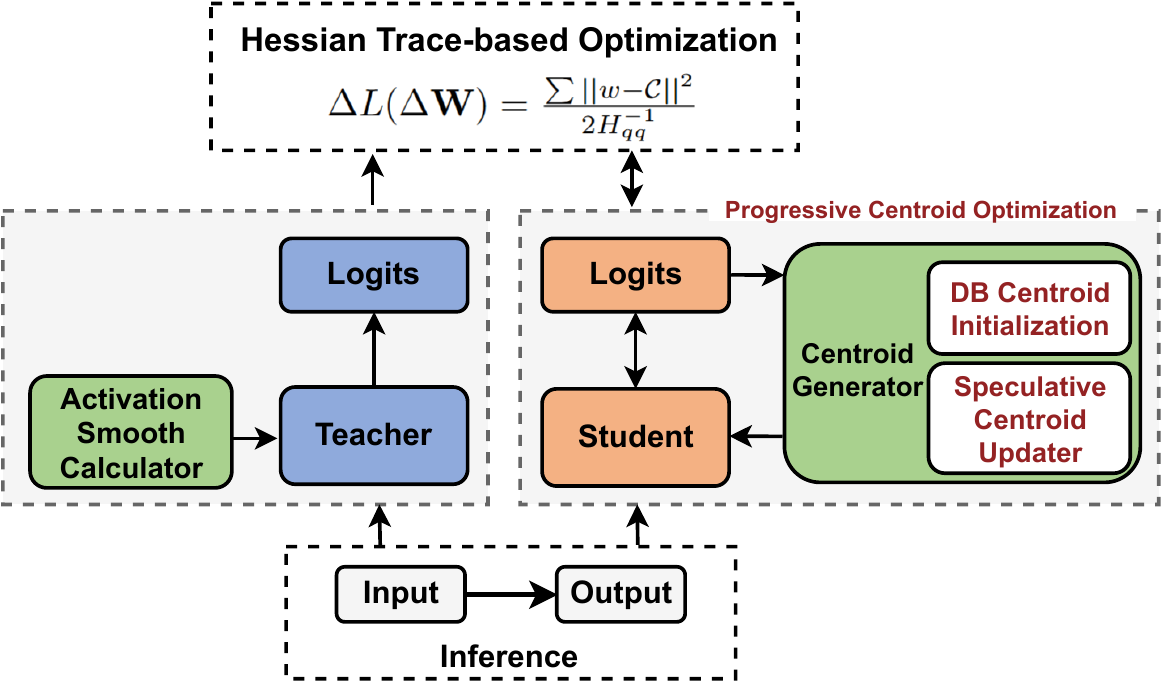}
\vspace{-10pt}
\caption{The LCD distillation framework.}
\label{fig:alg-framework}
\vspace{-10pt}
\end{figure}
In this section, we present \textbf{LCD}, a distillation-based framework for LLM weight clustering. As shown in Figure~\ref{fig:alg-framework}, the full-precision model serves as the teacher to guide the clustered student model. LCD employs a density-aware centroid initialization strategy to improve cluster quality and reduce the number of centroids efficiently. It further refines centroid positions and assignments using Hessian-guided objectives. To minimize accuracy degradation, LCD incorporates progressive and speculative optimization schemes that reduce centroid counts throughout training. Additionally, a smoothing mechanism is introduced to enable joint compression of weights and activations, achieving efficient end-to-end model compression.

\subsection{Density-Based Centroid Initialization}
\label{sec:alg-dbci}
Proper initialization of weight centroids is critical for reducing quantization errors and improving model accuracy. Here, LCD employs Density-Based Spatial Clustering of Applications with Noise (DBSCAN~\cite{schubert2017dbscan}) as the initialization scheme for centroids. By grouping densely packed data points based on their density, DBSCAN aligns well with diverse data distributions. Unlike traditional methods such as K-means, DBSCAN does not require predefining the number of clusters, providing greater flexibility and adaptability to varying datasets.
However, conventional DBSCAN~\cite{hanafi2022fast} requires carefully setting the \textbf{specified distance (eps)} and \textbf{minimum number of neighbors (MinPts)} to achieve effective clustering. To this end, we propose the \textbf{Density-Based Centroid Initialization (DBCI)}, designed specifically for the Gaussian-like weight distribution of LLMs, which often includes outliers. This algorithm derives parameter values directly from the data distribution. The process is outlined below:
\begin{enumerate}[itemsep=0pt, parsep=0pt, topsep=0pt]

    \item Sort the weights.
    \item Identify the parameters at 68.27\%, 95.44\%, and 99.74\% percentiles from the sorted positive and negative weights, corresponding to 1$\sigma$, 2$\sigma$, and 3$\sigma$ ranges. Calculate $\sigma$ as the mean of these six values:
    \begin{equation}
        \sigma = \frac{\scriptstyle w_{+\sigma} + w_{+2\sigma}+ w_{+3\sigma} + w_{-\sigma} + w_{-2\sigma} + w_{-3\sigma}}{12}
    \end{equation}
    \item Initialize all data points as unvisited. Select the two most extreme points as initial core points and gather all points within a $\sigma$-radius neighborhood to form two clusters.
    \item Count the points in both clusters, take the smaller count as MinPts, and calculate $\text{eps}$ as $\sigma / \text{MinPts}$.
    \item Apply the standard DBSCAN algorithm to the remaining unvisited points. Points with an eps-radius neighborhood containing at least MinPts are expanded into clusters; others are labeled as noise. Repeat until all points are clustered or marked as noise.
    \item For each cluster, compute the centroid $C$ that minimizes the L1-norm with all points in the cluster. These centroids form the initialization result.
\end{enumerate}
In such a way, DBCI efficiently generates a reliable and compact set of initial centroids without requiring prior knowledge. Empirical results show that DBCI reduces the number of initial weight centroids to 15–20, significantly decreasing the training time needed for subsequent optimization. Further details are provided in Section~\ref{sec:ablation}.

\subsection{Distillation with Hessian Matrix}
\label{sec:alg-hessian}

After initialization, given the original weights $\mathbf{W}$, we obtain the centroids $\mathbf{C}$, clustered weights $\mathbf{W}'$, and compression error $\Delta \mathbf{W} = \mathbf{W} - \mathbf{W}'$. The objective is to minimize the impact of quantization on the model's task performance, which translates to minimizing the expected deviation in the loss function $\mathcal{L}$:

\begin{equation}
\begin{aligned}
& \arg \min \quad \mathbb{E}[\mathcal{L}(\mathbf{X}, \mathbf{W}+\Delta \mathbf{W})-\mathcal{L}(\mathbf{X}, \mathbf{W})] \\
& \approx \Delta \mathbf{W}^T \cdot g(\mathbf{W})+\frac{1}{2} \Delta \mathbf{W}^T \cdot H(\mathbf{W}) \cdot \Delta \mathbf{W}
\end{aligned}
\end{equation}
where $g(\mathbf{W})$ is the gradient and $H(\mathbf{W})$ is the Hessian matrix. In this framework, the full-precision teacher model is assumed to have reached a local optimum, making the first-order term negligible ($g(\mathbf{W}) \approx 0$). Consequently, LCD leverages Hessian matrix optimization to achieve extreme clustering, and the problem reduces to:


\begin{equation}
\begin{aligned}
\arg \min _{\Delta \mathbf{W}} & \Delta \mathbf{W}^T \cdot H(\mathbf{W}) \cdot \Delta \mathbf{W} \\
& \text { s.t. } \quad \Delta \mathbf{W}=\mathbf{0}
\end{aligned}
\end{equation}

Using the Lagrange method, the above can be further transformed to minimize:
\begin{equation}
\Delta L(\Delta \textbf{W}')=\frac{\sum\|w-\mathcal{C}\|}{2 \mathbf{H}_{i i}^{-1}}
\label{opt target}
\end{equation}
where $H_{ii}$ represents the diagonal elements of the Hessian matrix. To address the computational overhead of a full Hessian matrix, we use a diagonal approximation of $H(\mathbf{W})$, simplifying the optimization process. This diagonal approximation allows for efficient computation of $\mathbf{W}'$, significantly reducing overhead while maintaining accuracy, leading to the following weight update formula:


\begin{equation}
\mathbf{W}_{t+1}=\mathbf{W}_t-\eta \frac{\nabla L}{diag(\textbf{H}')}
\end{equation}
where $\eta$ is the learning rate.

To mitigate the performance degradation caused by precision reduction, we incorporate KD into this optimization process. Notably, instead of using KL divergence as in traditional distillation, we utilize the metric in Equation~\ref{opt target} to update the weights, which significantly enhances computational efficiency. Moreover, since our update target is the centroid of each cluster, after updating the weights in each round of distillation, we also update the centroid values. This update consists of two parts: reclassification of weights and the update of the centroid values.

After the weight update, some weights may experience significant shifts. To handle this, we perform a reclassification operation. During the initialization phase, we calculate the centroid of each cluster and sort them in ascending order. For each cluster centroid $\mathcal{C}$, we compute the distance to its left and right neighboring centroids, dividing each by 2:



\begin{equation}
d_{\mathrm {left }}=\frac{\mathcal{C}_i-\mathcal{C}_{i-1}}{2}, \quad d_{\mathrm{right}}=\frac{\mathcal{C}_{i+1}-\mathcal{C}_i}{2}
\end{equation}

For edge centroids, only one distance is considered. For each weight $w$ in the weight matrix, if its update $\Delta w$ satisfies $\Delta w < d_{\text{left}}$ or $\Delta w > d_{\text{right}}$, we move $w$ from its current cluster to the left or right cluster.

After reclassification, we update the centroid value according to Equation~\ref{opt target} by summing the increments for all weight points assigned to the cluster as the centroid's offset:



\begin{equation}
\begin{aligned}
    \Delta \mathcal{C}_i &= \sum \Delta w_{\mathcal{C}_{i}} + \sum (\mathcal{C}_{i+1} - \mathcal{C}_{i}) + \Delta w_{\mathcal{C}_{i+1}} \\& + \sum (\mathcal{C}_{i-1} - \mathcal{C}_{i}) + \Delta w_{\mathcal{C}_{i-1}}
\end{aligned}
\end{equation}
Here, the first term represents the weights within this cluster, while the next two terms account for weights reclassified into this cluster. Through reclassification and centroid updates, LCD maintains the accuracy of the clustering process.









\subsection{Optimization of the number of clusters}
\label{sec:alg-clu-opt}
LCD leverages the Hessian matrix for distillation to maintain model performance under an extremely low number of clusters. However, the aforementioned update objective relies on a fixed number of clusters, which fails to fully explore the compression potential with fewer clusters. To achieve efficient and highly compressed clustering, LCD proposes two distillation optimization techniques:

\textbf{Progressive Centroid Optimization:}
During the distillation process, we utilize the diagonally approximated Hessian matrix to update weights.
Here, we sum the diagonal elements and use the Hessian Trace for centroid optimization. Based on its characteristics, when its value approaches a predefined near-zero threshold $\theta$, it indicates that the current number of centroids can almost perfectly approximate the original weight distribution. Then, we will try to reduce the number of clusters. After the centroid update, we select the two closest centroids and merge them into one. The new centroid is determined as the weighted average of the original two centroids, where the weights are proportional to the number of points $n$ in each cluster:

\begin{equation}
    \mathcal{C}_{new} = \frac{n_b \mathcal{C}_a + n_a\mathcal{C}_b}{n_a+n_b}
\end{equation}

\textbf{Speculative Centroid Optimization:}
While progressive centroid optimization ensures stable clustering, it may fall into a local optimum due to initialization, as it produces minimal changes in optimization results. To address this, we introduce a speculative centroid search to complement the progressive optimization framework.

When the progressive search stabilizes without reducing the number of centroids, and the Hessian trace no longer changes monotonically, we regard the clustering as a local optimum. At this point, we double the eps-radius used during initialization, reinitialize the centroids, and perform iterative optimization for a fixed number of iterations, $p$. The goal is to achieve a clustering result that meets the inference accuracy threshold, $\Theta$. If successful within $p$ iterations, the process resumes with progressive centroid optimization based on the new centroids. If unsuccessful, the framework reverts to the original clustering result, reduces eps from 2eps to 1.5eps, and initiates another round of speculative search. This process repeats until a predefined training round limit, $T$, is reached or the clustering result stabilizes. The final solution is the lowest number of centroids that yields the optimal result.

Through progressive and speculative centroid optimization, LCD effectively distills the weight matrix, achieving an extremely low number of quantized centroids while maintaining high accuracy.

\subsection{Adaptive Smooth Optimization}
\label{sec:smooth-opt}

While clustering activations is feasible for complex data distributions, fixed centroid values can lead to unstable performance due to variations in input data distributions. Dynamic clustering of inputs, however, introduces significant computational costs and latency during inference. Prior efforts have shown that common quantization methods fail to achieve effective compression, as 8/16 bits are necessary to maintain model performance due to the presence of outliers.

\begin{figure}[t!] 
\vspace{-5pt}
\setlength{\abovecaptionskip}{3pt}
\setlength{\belowcaptionskip}{3pt}
\centering
\includegraphics[width=1.0 \linewidth]{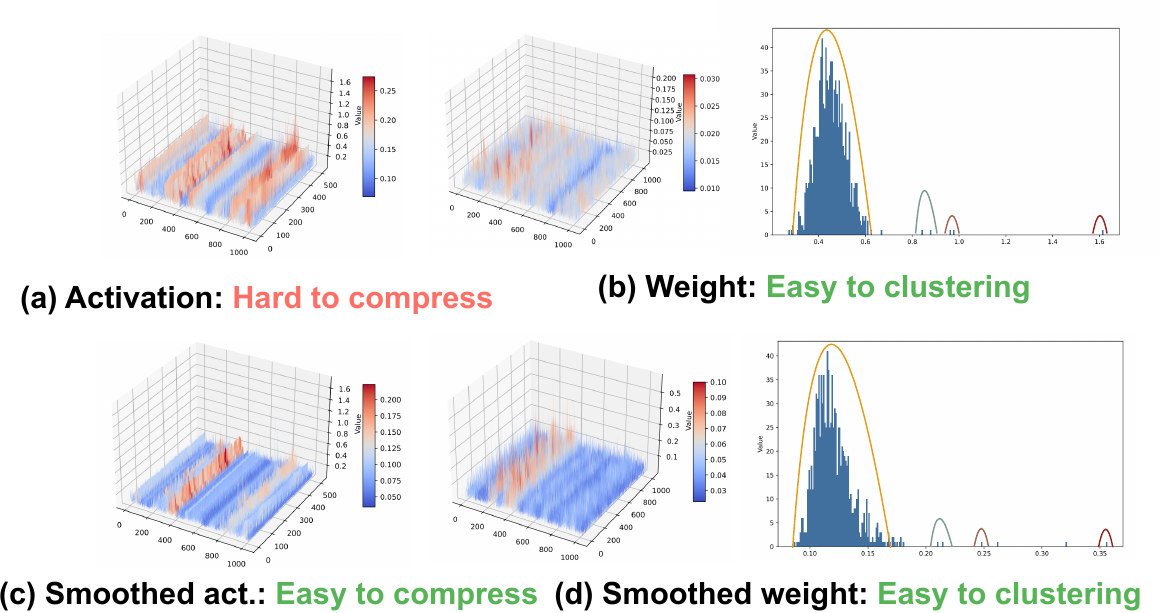}
\vspace{-15pt}
\caption{After smoothing, activations are easier to quantize, while the difficulty of clustering the smoothed weights is stable.}
\label{fig:alg-smooth}
\end{figure}

An effective solution is to use smoothing. By smoothing the activations, we make the input more quantization-friendly, enabling higher compression rates. The trade-off is that this increases the complexity of the weight data distribution. However, as demonstrated in previous work and shown in Figure~\ref{fig:alg-smooth}, clustering exhibits robust compression performance across various parameter distributions.


To achieve better compression, we propose combining input smoothing with weight clustering. By applying smoothing to reduce the bit-width required for activations, while maintaining weight clustering performance with minimal degradation, this approach enables efficient, synergistic compression of both activations and weights.
Specifically, we adopt a layer-wise fixed smoothing parameter. Using a calibration dataset, we run inference on the teacher model and compute the mean squared error (MSE) between activations before and after smoothing, both following INT8 quantization. The smoothing parameter $s_m$ that minimizes this MSE is selected as the fixed smoothing factor for each layer. The optimization objective is:




\begin{equation}
    \underset{s_m}{min} \quad  MSE(X, Q_{INT8}(X/s_m)\times s_m))
\end{equation}


After adaptively determining the smoothing parameters, we scale the weights of each layer in the full-precision teacher model by the corresponding smoothing factor. The resulting smoothed weights are then used during the distillation process for the student model. During inference, inputs are divided by the same smoothing factors to maintain consistency. Formally, we transform the weight matrix $\mathbf{W}$ to $Smooth(\mathbf{W})$, and correspondingly adjust the inference procedure.
Applying smoothing enables quantization of activations at lower bit-widths, thereby achieving more effective compression. Crucially, this method allows integer quantization of activations without altering the number of weight clustering centroids, preserving model accuracy while further improving computational efficiency.







\section{Inference with Table Lookup}
\label{sec:infer design}
\begin{figure}[t!] 
\vspace{-5pt}

\setlength{\abovecaptionskip}{3pt}
\setlength{\belowcaptionskip}{3pt}
\centering
\includegraphics[width=0.75 \linewidth]{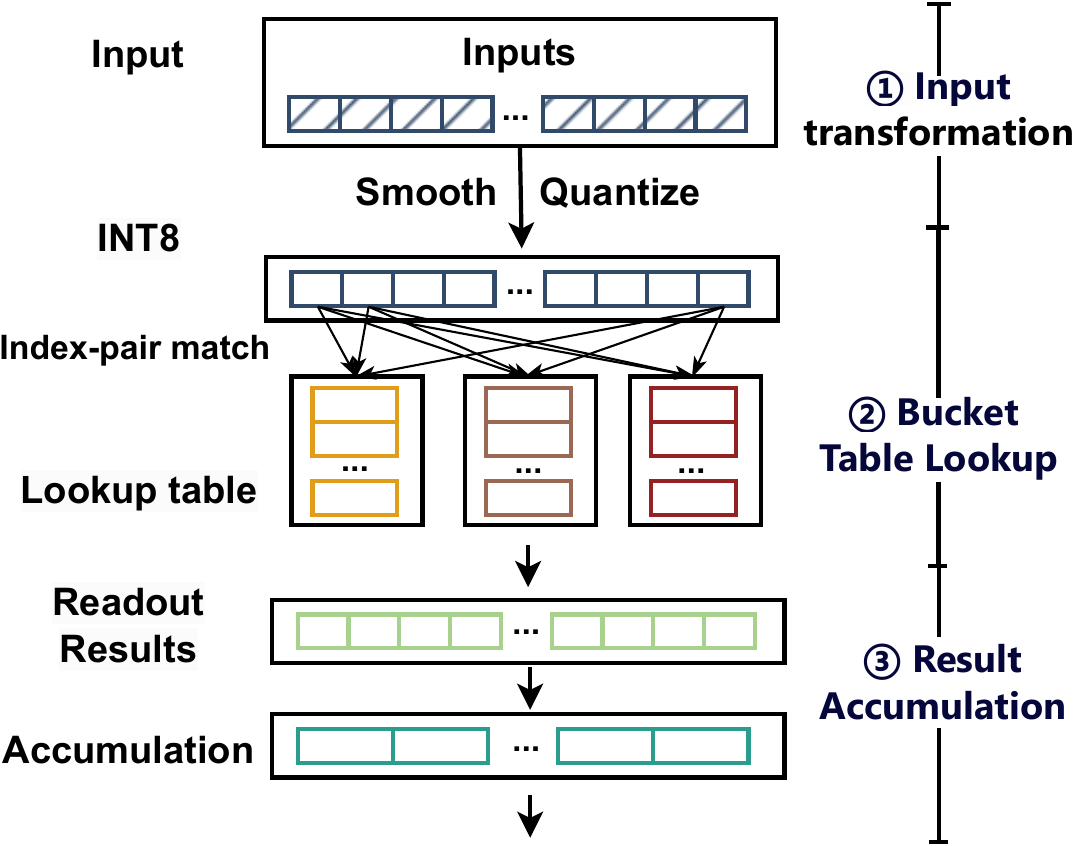}
\vspace{-3pt}
\caption{Depiction of LCD table lookup inference system.}
\label{fig:alg-infer}
\vspace{-15pt}
\end{figure}

In this section, we present the LCD inference design, illustrated in Figure~\ref{fig:alg-infer}, which comprises three stages: Input Transformation, Bucket Table Lookup, and Accumulation. The input tensor is first smoothed and quantized into low bit-width integer indices. Then, pre-computed values are retrieved from lookup tables and accumulated. Leveraging hardware optimizations like T-MAC~\cite{wei2024tmaccpurenaissancetable} and LUT Tensor Cores~\cite{mo2024lut}, this design replaces floating-point multiplications with efficient table lookups, eliminating multiplications and reducing centroid addressing overhead during inference.

\subsection{Input Transformation Stage.} 

In the input transformation stage, we aim to efficiently convert floating-point inputs into low-bit-width indices for table lookup. This involves smoothing and quantizing the input, using pre-determined parameters for each layer. These parameters are computed offline using a calibration set, and the corresponding LUTs are constructed ahead of time. The smoothing transformation is applied layer-wise, as discussed in Section~\ref{sec:smooth-opt}, while quantization employs uniform symmetric quantization to maximize computational efficiency. For an input $\mathbf{X}$, the quantized value is computed as:
\begin{equation}
    q = Clip\lfloor\frac{X_m}{s_q}\rceil , q\in[-2^{b},2^{b}-1]
\end{equation}
Here, $b$ represents the bit-width, which is set to 8. Since the smoothing transformation is followed by the quantization step, these two operations can be combined into one:

\begin{equation}
    q = Clip\lfloor\frac{X}{s_ms_q}\rceil , q\in[-2^{b},2^b-1]
\end{equation}
This reduces the computation to a single multiplication, equivalent to multiplying by $1/s_m s_q$). After rounding, the input is converted into an integer index for the lookup. By storing these fixed parameters in registers, we can efficiently transform the input into the required index with minimal storage and computational overhead.




\subsection{Bucket Table Lookup and Accumulation.}
After the input transformation, the input index is used in the bucket table lookup and accumulation stage to compute the final results. In this stage, we read pre-computed values from a lookup table based on the input index and then sum (accumulate) these values to get the final result. By using the lookup table, LCD avoids complex multiplications, making the process faster and reducing computational costs.

However, using lookup tables introduces challenges for inference efficiency. First, table read can be slow because it's hard to parallelize and introduces extra indirect memory
accesses, which exaggerate memory overhead. Second, as the table size grows, the time and memory needed for table lookups become more pronounced.


To this end, LCD uses a centroid-stationary bucket LUT strategy, dividing the large lookup table into smaller buckets based on weight centroids. Each bucket stores precomputed values for its centroid group, reducing table size and enabling parallel lookups for faster inference.
By leveraging symmetric quantization, we store results only for non-negative input indices and apply sign adjustments during accumulation.

LCD’s distillation reduces centroids to fewer than 16, allowing compact 4-bit representation that lowers memory usage and boosts speed compared to traditional LUTs. Additionally, weight smoothing removes the need for dequantization during lookup, enabling direct accumulation and maintaining system simplicity and efficiency.

\section{Evaluation}

\subsection{Experimental Settings}

\textbf{Benchmark.} We benchmark LLaMA-2~\cite{touvron2023llama} on general language tasks, report perplexity on language modeling tasks, including WikiText-2~\cite{merity2016pointer} and C4~\cite{raffel2020exploring}, and perform zero-shot evaluations on common sense QA benchmarks, including PIQA~\cite{bisk2020piqa}, HellaSwag~\cite{zellers2019hellaswag}, WinoGrande~\cite{sakaguchi2021winogrande} and ARC~\cite{Clark2018ThinkYH}. To validate the generality of the LCD scheme, we also  evaluate Bert-large~\cite{kenton2019bert} with GLUE~\cite{wang2018glue} and GPT2-XL~\cite{radford2019language} with WikiText-2. We exploit pre-trained network
models from TorchVision and Huggingface Model Zoo, and their validation accuracy with FP16 as baseline.
And, when training Bert, our calibration data consists of 128 random samples of the GLUE-SST-2 dataset, and 128 random segments of the WikiText-2 for GPT and LLaMA.

\textbf{Baselines.} The accuracy baseline contains various methods, including:1) FP16, we use it as a basis. 2) QServe~\cite{lin2024qserve}, an algorithm and system co-design framework that performs progressive quantization during computation; 3) GPTQ~\cite{frantar2022gptq}, a weight PTQ  method based on approximate second-order information; 4) QuIP\#~\cite{tseng2024quip}, a weight PTQ method using randomized Hadamard transform and vector quantization; 5) LLM-QAT~\cite{liu2023llm}, a QAT data-free distillation method; 6) BitDistiller~\cite{du2024bitdistiller}, a framework that synergizes QAT with KD framework using asymmetric quantization and clipping. 7) SKIM~\cite{bai2024skim}, a scaled K-means clustering with mixed precision. 
For inference performance, the baselines include  QServe and TVM~\cite{chen2018tvm}, two mainstream neural network acceleration tools, as well as LUT-NN~\cite{tang2023lut}, an inference acceleration method based on centroid learning and table lookup. LCD adopts the inference framework described in Section~\ref{sec:infer design}. For models not provided in their original papers, we reproduce their experiments and report the results. All experiments are conducted on A100 GPUs.

\subsection{Evaluation Results}

\begin{table}[]
\centering
\renewcommand{\arraystretch}{1}
\resizebox{\linewidth}{!}{
\begin{tabular}{c|c|cc}
\toprule[1.5pt]
Model           & Bert-large  & \multicolumn{1}{c|}{GPT2-XL} & LLaMA-2-7B \\ \hline
Dataset         & SST-2(Acc\%)$\uparrow$  & \multicolumn{2}{c}{Wikitext-2(PPL)$\downarrow$}         \\ \midrule \midrule
Baseline        & 92.9        & \multicolumn{1}{c|}{18.34}    & 5.47       \\ \hline
LCD performance & 92.7        & \multicolumn{1}{c|}{18.78}    & 5.77       \\ \hline
Centroids       & 5           & \multicolumn{1}{c|}{6}        & 8          \\ \bottomrule[1.5pt]
\end{tabular}
}
\vspace{-5pt}
\caption{Comparison of accuracy and clustering performance.}
\label{tab:centroid-result}
\vspace{-0.3cm}
\end{table}

\begin{table*}[]
\vspace{-0.6cm}
\footnotesize
\centering
\renewcommand{\arraystretch}{0.95}
\begin{tabularx}{\textwidth}{c||>{\centering\arraybackslash}X
|>{\centering\arraybackslash}X|>{\centering\arraybackslash}X|>{\centering\arraybackslash}X|>{\centering\arraybackslash}X|>{\centering\arraybackslash}X|>{\centering\arraybackslash}X}
\toprule[1.5pt]
\textbf{LLaMA-2-7B} & bits(\#C)$\downarrow$ & Wikitext2$\downarrow$ & C4$\downarrow$   & PIQA$\uparrow$  & Hella.$\uparrow$  & Wino.$\uparrow$  & ARC-c$\uparrow$  \\ \midrule \midrule
FP 16               & 16       & 5.47      & 6.97 & 78.4  & 57.1   & 68.4   & 43.3  \\ \hline
QServe~\cite{lin2024qserve}             & 4        & 6.76      &  -    & 78.1  & 56.0   & 68.6   & 44.8 \\ \hline
GPTQ~\cite{frantar2022gptq}                & 3        & 6.38      &   -   & 75.5  & 51.7   & 67.2   & 38.4 \\ \hline
LLM-QAT~\cite{liu2023llm}             & 3        & 6.02      &  -    & 77.3  & 54.7   & 68.4   & 40.6 \\ \hline
QuIP\#~\cite{tseng2024quip}              & 3        & 5.79      & 7.32 & 77.3  & -      & 66.5   & 39.2  \\ \hline
BitDistiller~\cite{du2024bitdistiller}        & 3        & 5.97      & -    & 77.0 & 55.4   & 68.4   & 41.2  \\ \hline
\multirow{2}{*}{SKIM~\cite{bai2024skim}} & 3.2*     & 6.07      & 7.52 & -     & -      & -      & -     \\ \cline{2-8} 
                      & 3*       & 6.21      & 7.68 & -     & -      & -      & -     \\ \hline
                       & 3.3*(10)              & \textbf{5.58}                  & \textbf{7.25}           & \textbf{78.1}           & \textbf{56.5}             & 68.3            & 42.0            \\ \cline{2-8} 
\multirow{-2}{*}{Ours} & 3*(8)                 & \textbf{5.77}                  & \textbf{7.42}           & 77.3           & 55.8             & 68.1            & 41.6            \\ \bottomrule[1.5pt]
\end{tabularx}
\vspace{-5pt}
\caption{Wikitext2, C4 perplexity and accuracy on five Commonsense QA tasks for LLaMa-2 7B. 
Avg. bit marked with $*$ indicate the equivalent bitwidth. In this table, 
Hella. is abbreviation for Hellaswag, Wino. for Winogrande and ARC-c for Arc-challenge.
}
\label{tab:llama-result}
\vspace{-12pt}
\end{table*}

\begin{figure}[t!] 
\vspace{-5pt}
\setlength{\abovecaptionskip}{3pt}
\setlength{\belowcaptionskip}{3pt}
\centering
\includegraphics[width=0.8 \linewidth]{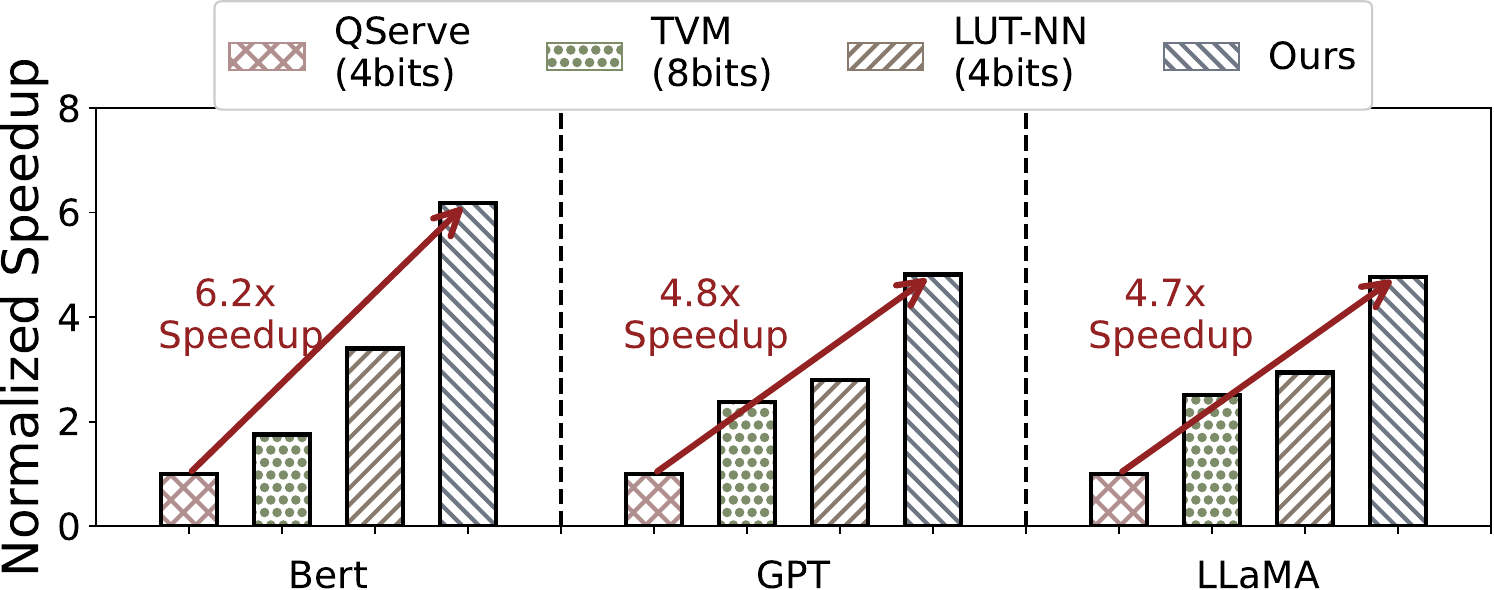}
\vspace{-3pt}
\caption{Performance comparison across networks: 2.3 bits, 2.6 bits, and 3 bits correspond to 5, 6, and 8 centroids, respectively.}
\vspace{-15pt}
\label{fig:speedup}
\end{figure}

\textbf{Performance.}
The results in Table~\ref{tab:centroid-result} show that weights can be effectively clustered into a small number of centroids, achieving compression equivalent to 2-3 bits. Remarkably, this level of compression maintains accuracy comparable to the original models. 
Table~\ref{tab:llama-result} compares LCD’s performance with previous methods on general language tasks. LCD outperforms competing methods in terms of perplexity on WikiText-2 and C4, while also delivering consistent results across various QA benchmarks. Compared to SKIM, which also performs clustering, LCD shows superior performance. Notably, LCD achieves activation compression with extremely low centroids, a capability not found in other PTQ/QAT/clustering methods. This gives LCD a distinct advantage, highlighting its effectiveness and scalability.

\textbf{Speedup.} Figure~\ref{fig:speedup} illustrates the end-to-end speedup of LCD compared to Qserve and other methods. LCD achieves speedups of 6.2x, 4.8x, and 4.7x on BERT-large, GPT2-XL, and LLaMA-2-7B, respectively. Although LLaMA, with its larger input tensor size, achieves significant performance gains after quantization, the increase in the average number of centroids reduces lookup table efficiency, partially offsetting the speedup. 


\subsection{Ablation Study}
\label{sec:ablation}
\begin{figure}[t!]
    \centering
    \begin{subfigure}[b]{0.45\linewidth}
        \centering
        \includegraphics[width=\textwidth]{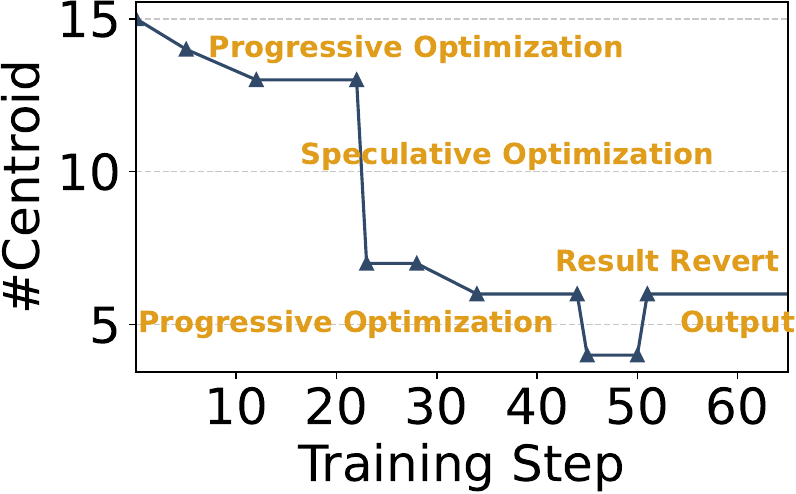}
        \caption{Centroid changes during the LCD distillation process.}
        \label{fig:aba-single cluster}
    \end{subfigure}
    \hfill
    \begin{subfigure}[b]{0.45\linewidth}
        \centering
        \includegraphics[width=\textwidth]{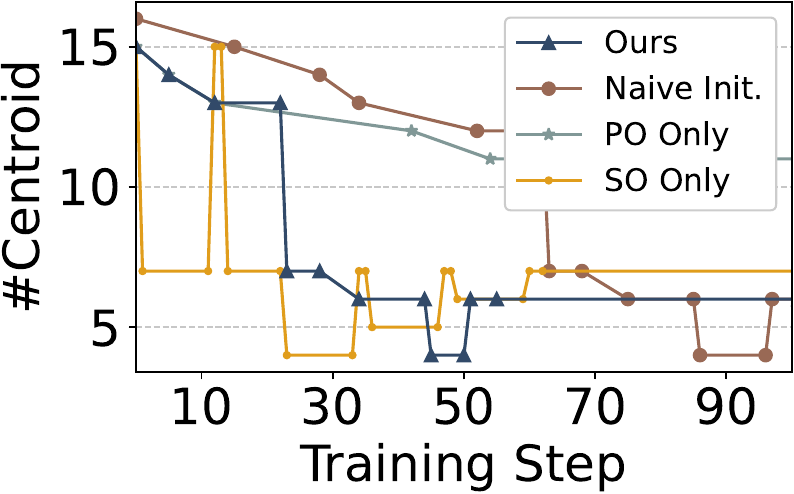}
        \caption{Centroid changes during diverse distillation processes.}
        \label{fig:aba-cluster compare}
    \end{subfigure}
    \vspace{-10pt}
    \caption{Centroid Change vs. Training Steps (GPT2-XL).}
    \label{fig:aba-cluster}
    \vspace{-10pt}
\end{figure}

\textbf{Distillation Optimization Techniques.}
LCD uses DBCI centroid initialization combined with progressive and speculative optimizations for efficient clustering. Starting from 15 centroids initialized by DBCI (Figure~\ref{fig:aba-single cluster}), progressive optimization quickly reduces centroids. Once stability and Hessian Trace fluctuations are detected, speculative optimization further cuts centroids to 7. Afterward, progressive optimization resumes, converging finally to 6 centroids as output.
We compare three strategies: naive 4-bit centroid initialization (``Naive init.''), progressive optimization only (``PO only''), and speculative optimization only (``SO only''). Figure~\ref{fig:aba-cluster compare} shows that naive initialization needs more steps due to lower accuracy, progressive-only converges prematurely at 11 centroids, and speculative-only yields unstable, suboptimal results.

\textbf{Activation Smooth.} 
LCD applies adaptive smoothing to activations to reduce quantization errors and enable higher compression. Table~\ref{tab:aba-smooth} summarizes model performance under different smoothing levels after activation quantization, along with the resulting centroid counts from weight clustering. Results show that smoothing improves quantization accuracy significantly. Although increasing smoothing compresses activations more, it makes weight clustering harder, increasing centroids. LCD’s adaptive smoothing outperforms fixed smoothing by dynamically calibrating each layer offline, achieving optimal quantization without adding inference overhead.

\begin{table}[]
\centering
\renewcommand{\arraystretch}{1.5}
\resizebox{\linewidth}{!}{
\begin{tabular}{c|cc|cc|cc|cc}
\toprule[1.5pt]
                  & \multicolumn{2}{c|}{Origin}      & \multicolumn{2}{c|}{$S_m$ = 0.5} & \multicolumn{2}{c|}{$S_m$ = 0.8} & \multicolumn{2}{c}{Ada Smooth (Ours)}    \\ \midrule[1.5pt]
Activation format & \multicolumn{1}{c|}{FP16} & INT8 & \multicolumn{1}{c|}{INT8} & INT4  & \multicolumn{1}{c|}{INT8}  & INT4 & \multicolumn{1}{c|}{INT8} & INT4  \\ \hline
PPL               & \multicolumn{1}{c|}{5.75} & 56.2 & \multicolumn{1}{c|}{5.92} & 324.5 & \multicolumn{1}{c|}{5.68}  & 8.42 & \multicolumn{1}{c|}{5.77} & 10.25 \\ \hline
\#Centroid         & \multicolumn{1}{c|}{8}    & 8    & \multicolumn{1}{c|}{8}    & 8     & \multicolumn{1}{c|}{14}    & 14   & \multicolumn{1}{c|}{8}    & 8     \\ \bottomrule[1.5pt]
\end{tabular}
}
\caption{Analyzation of smoothing settings (LLaMa-2 7B).}
\label{tab:aba-smooth}
\end{table}

\begin{figure}[t!] 
\vspace{-7pt}
\setlength{\abovecaptionskip}{3pt}
\setlength{\belowcaptionskip}{3pt}
\centering
\includegraphics[width=0.8 \linewidth]{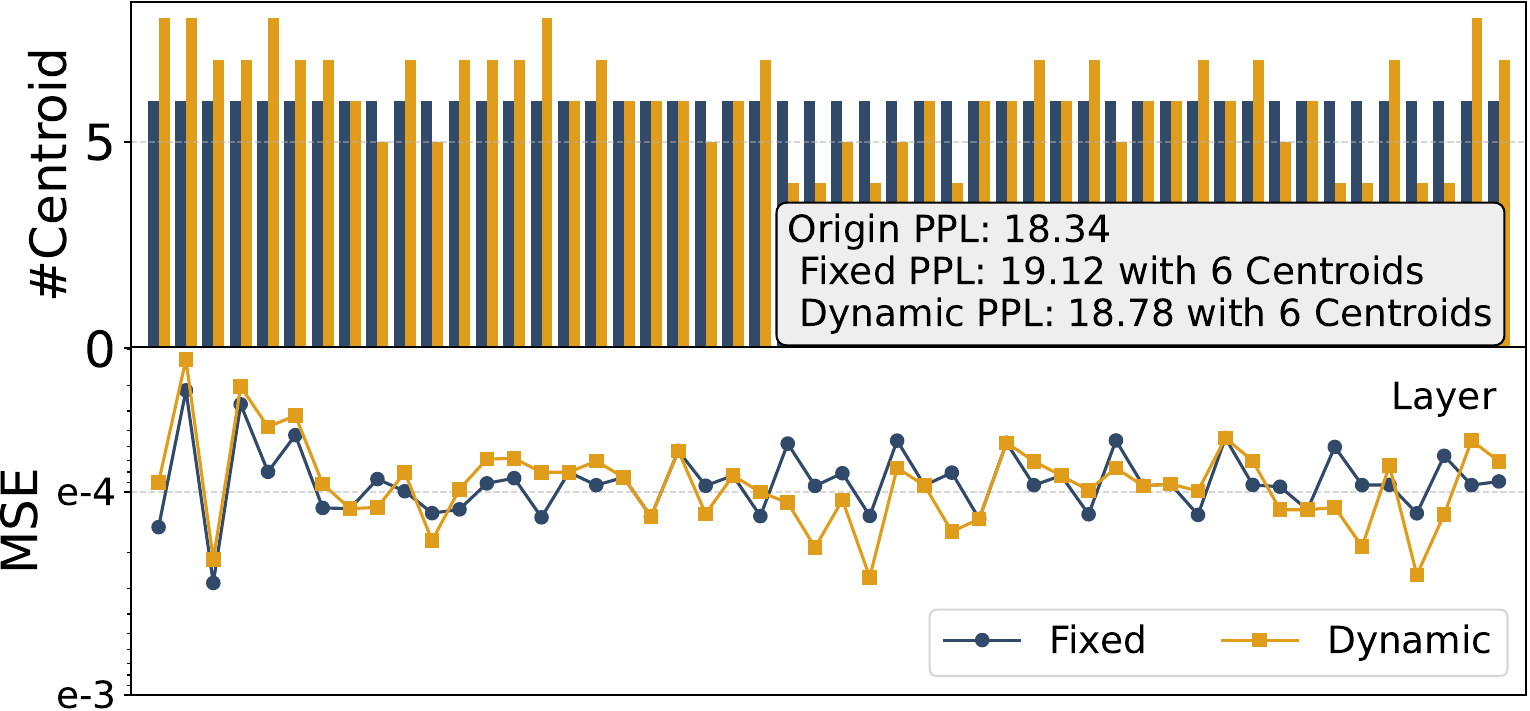}
\vspace{-2pt}
\caption{Layer-wise Centroids and MSE for fixed and dynamic clustering strategy on GPT2-XL.}
\label{fig:bitwidth}
\vspace{-8pt}
\end{figure}

\textbf{Layer-wise Dynamic Centroids.}
Given the variability in weight distributions across layers, optimized centroids vary per layer. LCD dynamically adjusts centroid counts layer-wise and reports the average number, as shown in Figure~\ref{fig:bitwidth}. The results indicate earlier layers keep more centroids, while LCD’s dynamic allocation maintains an average of 6 centroids, achieving superior performance.

\section{Conclusion}

In this paper, we present LCD, a novel framework that tackles the challenges of deploying large language models by enabling ultra-low bit compression without sacrificing performance. By integrating clustering with knowledge distillation and leveraging advanced optimization techniques, LCD reduces weight bitwidths to 2–3 bits effectively. Furthermore, LCD employs an LUT-based design to significantly accelerate inference. Evaluations show that LCD outperforms existing methods, achieving remarkable speedups during inference.

\clearpage

\section*{Limitation}
Although LCD can effectively cluster weights into low-bit representations, the training process is quite time-consuming. In particular, when LCD introduces adaptive smoothing, the adaptive determination of the optimal smoothing coefficient significantly expands the search space, rendering the training process of our approach relatively inefficient.

\bibliography{ref}




\end{document}